\documentclass[conference]{IEEEtran}
\IEEEoverridecommandlockouts
\usepackage{graphicx, subfigure}
\usepackage{amssymb}
\usepackage{amsmath}
\usepackage{url}
\usepackage{hyperref}
\usepackage{collectbox}
\usepackage{cite}
\def\BibTeX{{\rm B\kern-.05em{\sc i\kern-.025em b}\kern-.08em
    T\kern-.1667em\lower.7ex\hbox{E}\kern-.125emX}}
\begin{document}

\title{Object cosegmentation using deep Siamese network}

\author{\IEEEauthorblockN{Prerana Mukherjee\IEEEauthorrefmark{1},
Brejesh Lall\IEEEauthorrefmark{1}  and
Snehith Lattupally\IEEEauthorrefmark{1}
}
\IEEEauthorblockA{\IEEEauthorrefmark{1}Dept of EE, IIT Delhi, India. \\Email: \{eez138300, brejesh, eet152695  \}@ee.iitd.ac.in} }

% \and
% \and
% \and
% \and
% \and

\maketitle

\begin{abstract}
Object cosegmentation addresses the problem of discovering similar objects from multiple images and segmenting them as foreground simultaneously. In this paper, we propose a novel end-to-end pipeline to segment the similar objects simultaneously from relevant set of images using supervised learning via deep-learning framework. We experiment with multiple set of object proposal generation techniques and perform extensive numerical evaluations by training the Siamese network with generated object proposals. Similar objects proposals for the test images are retrieved using the ANNOY (Approximate Nearest Neighbor) library and deep semantic segmentation is performed on them. Finally, we form a collage from the segmented similar objects based on the relative importance of the objects. 
\end{abstract}

\begin{IEEEkeywords}
Cosegmentation, Siamese Network, Multiscale Combinatorial Grouping, Nearest Neighbor
\end{IEEEkeywords}

\section{Introduction \label{sec:intro}}
Automated foreground segregation and localization of objects constitute the fundamental problem in computer vision tasks. Further the lack of sufficient information about the foreground objects makes it highly complex to deal with it. The exploitation of the commonness prior and the joint processing of similar images (containing objects of same category) can aid in the process of such object related tasks. Cosegmentation refers to such class of problems which deals with the segmentation of the \textit{common objects} from a given set of images without any priori knowledge about the foreground. It was first hypothesized in \cite{vicente2011object} that in most cases the common objects for cosegmentation represent the `objects of interest' which appear in the images instead of common background details. These objects have huge variations in terms of scale, viewpoint, rotation, illumination, location and affine changes. In other cases, it may be highly occluded by other objects. Even same class of objects may drastically differ in appearance resulting in high intra-class variation. 

The works in \cite{wang2015robust, li2016object, Huang2016ObjectCB} solve the generic object cosegmentation by applying the localization constraint that in all the images the common object will always belong to the salient region. Some of the methods are confined to the cosegmentation between image pairs \cite{rother2006cosegmentation, wang2013image, joulin2012multi} while others require some user intervention \cite{batra2010icoseg, kowdle2010imodel}. Further \cite{rother2006cosegmentation, mukherjee2009half} pose it as segmenting only those objects that are exactly similar. These approaches are unable to handle the intra-class variations or other synthetic changes or noise which might be present in case of images that are downloaded from Internet. In recent years, with increase in the computational power and access to widespread availability
of semantic annotations for object classes, deep learning has achieved dramatic break-through in various applications. Siamese Network has also been extensively used for many vision applications. They have been used to learn the similarity metrics by aligning the similar objects together and dissimilar objects far away. This motivates us to solve the cosegmentation problem using the high-level features extracted using deep networks. We propose to couple the similarity based clustering and cosegmentation task so that they can coexist and benefit from each other synergistically.

In this work, we pose cosegmentation as a clustering problem using the Siamese network. For a given set of images, we train the Siamese twin architecture to assess the similarity of two equally sized patches. These patches are the object proposals of an image. Co-segmenting the objects using trained model is done using high-level features utilizing  fully convolutional networks \cite{long2015fully} rather than low-level features like SIFT, HOG etc. Finally, we create a visual summary from the segmented images based on their similarity score in the respective class. In view of the above discussions, the major contributions of this paper are:

\begin{enumerate}
\item Cosegmentation is posed as a clustering problem to align the similar objects using Siamese network and segmenting them. We also train the Siamese network on non-target classes with no to little fine-tuning 
and test the generalization capability to target classes.
\item Generation of visual summary of similar images based on relative relevance.
\end{enumerate}

Rest of the paper is organized as follows. 
%In Section \ref{sec:related} we discuss the related works. 
In Sec. \ref{sec:Methodology}, we describe the proposed approach in detail. In Sec. \ref{sec:results}, we present the results and discussions. Finally, we conclude the paper in Sec. \ref{sec:conclusion}.

% \section{Related Work}
% \label{sec:related}
% In recent years, a number of methods have been developed for cosegementation problem. Pioneering work was done by Rother et al.\cite{rother2006cosegmentation} who formulated the object cosegmentation of image pairs as histogram matching problem and then incorporated a constraint (global) into Markov Random Fields (MRFs). Recent works in cosegmentation can be categorized into three categories: i) Saliency Based, ii) Graph Based and iii) Clustering Based approaches. Yong Li et al.\cite{li2016object} formulated cosegmentation using Adaptive Discriminative low rank matrix discovery (ADLRR) to find the common and salient regions in each image. Rubinstein et al.\cite{Rubinstein13Unsupervised} formulated the cosegmentation as object discovery and joint segmentation problem.% Koteswar rao et al.\cite{jerripothula2016image} made use of inter-image information using co-saliency and later performed single-image segmentation for each of the individual image. 
% Lehu Huang et al.\cite{Huang2016ObjectCB} formulated the cosegmentation as the combination of saliency and propagation of superpixel similarity. Graph based methods \cite{meng2012object, li2016unsupervised} have been exploited as shortest path problem, graph labeling to cosegment similar images. Initial clustering of images based on feature statistics reduce the segmentation to a subset of common groups of images \cite{li2014unsupervised, Lucas15cseg}. 

\section{Methodology} \label{sec:Methodology}
\begin{figure*}[ht]
\centering
\fbox{
\includegraphics[scale=0.75]{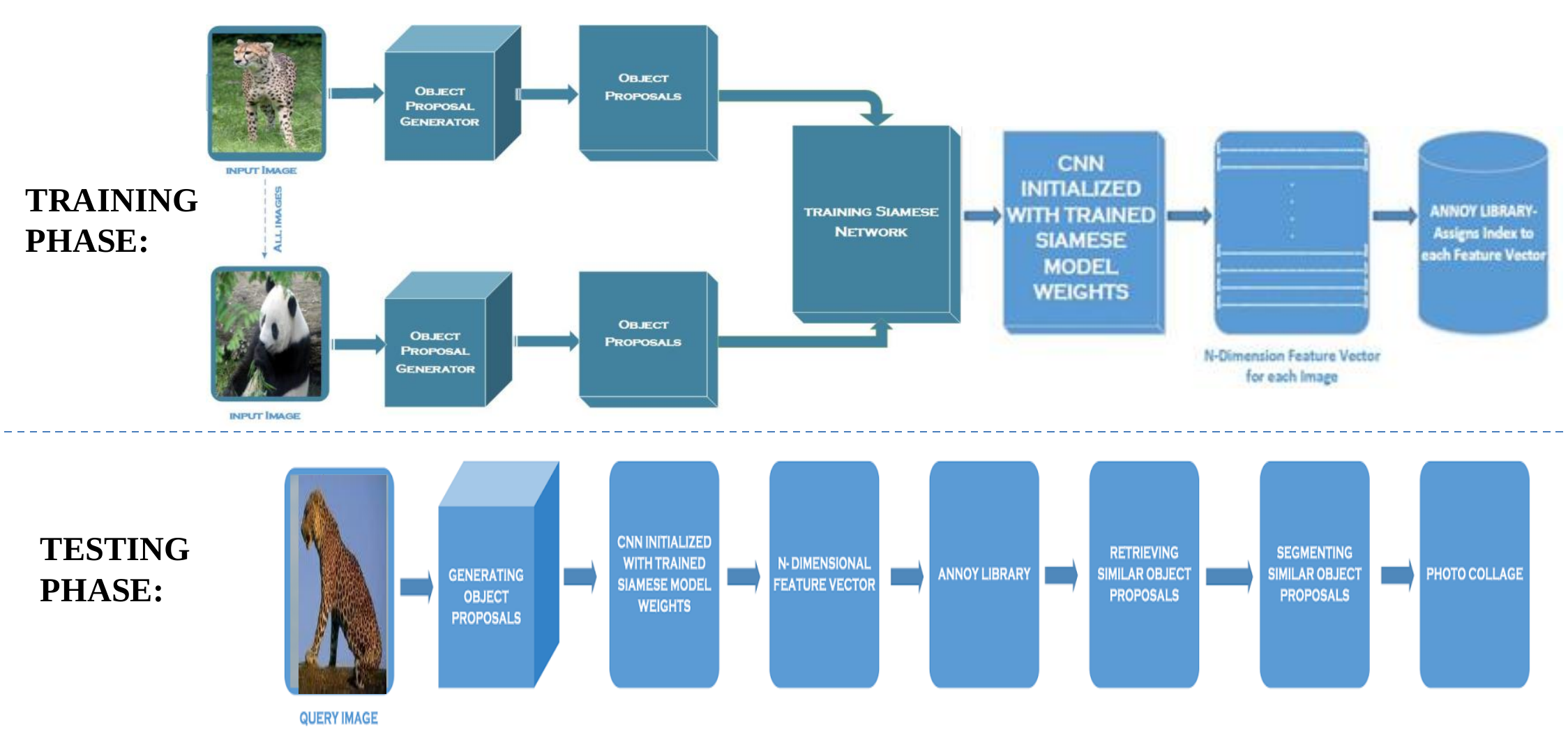}}
 % 106.png: 729x567 pixel, 72dpi, 25.72x20.00 cm, bb=0 0 729 567
\caption{Overall Architecture}
\label{fig:block-diagram}
\end{figure*}

In the following subsections, we describe the components of the proposed method. Fig. \ref{fig:block-diagram} shows the overall pipeline of the proposed method.
\subsection{Siamese network}
For given set of images, we first generate the object proposals using different object proposal techniques as described in Sec. \ref{sec:results}. The generated object proposals are given to Siamese Network for training. Siamese Networks are useful in finding similarities and relationship between different structures. The Siamese configuration consists of two convolutional neural networks (CNNs) with shared weights with a contrastive loss layer. 
The input to the Siamese network are two input patches (object proposals) along with a similarity label. Similarity label `1' indicates that patches are similar while `0' indicates dissimilar patches. Two CNNs generate a N-Dimensional feature vector in forward pass. The N-Dimensional vectors are fed to the contrastive loss layer which helps in adjusting the weights such that positive samples are closer and negative samples are far from each other. Contrastive loss function penalizes the positive samples that are far away and negative samples that are closer. Let us consider two  patches $(x_1,x_2)$ that are fed to Siamese network. Let us assume the N-Dimension vectors generated by convnets as $f(x_1)$ and $f(x_2)$. $Y$ be the binary label $Y\epsilon\{0,1\}$, $Y$=1 for similar pairs and 0 otherwise. Margin $m$ is defined for the contrastive layer such that positive samples are at a distance less than margin and negative samples are at a distance greater than margin. Thus, the contrastive loss function is given as,
\begin{equation}
     \textit{L}(W, Y, x_1,x_2)=Y\frac{1}{2} D_W^2 + (1-Y) \frac{1}{2}\{max(0,m-D_W^2)\}
\end{equation} 
where $D_W^2=\left \| f(x_1)-f(x_2) \right \|^2$ is the Euclidean distance between the two feature vectors of the input patches. The outputs from the fully-connected layers are fed to contrastive layers, which measures the distance between two features. The weights $W$ are adjusted such that the loss function is minimized.
%We used stochastic gradient descent with momentum to adjust the weights. %The Softmax layer and Accuracy layers are removed and an inner product layer with output equal to N-Dimension is appended. 

After training the Siamese Network, we deployed the trained model on test images. First we extracted the object proposals for the test images. A N-Dimensional feature vector is generated for each of the proposals. In our experiments, we used 256-Dimensional feature vector. The features generated for test image proposals using trained Siamese network are fed to Annoy (Approximate Nearest Neighbor) Library \footnote{https://github.com/spotify/annoy}. It measures the Euclidean distance or Cosine distance between vectors. It works by building up a tree using random projections. A hyper-plane is generated at every intermediate node in the tree which divides the space into two parts. The hyperplane is chosen such that it is equidistant from the chosen two sample points. Annoy library allows to tune two parameters, number of trees and number of nodes to be checked while searching. The features extracted from the test image proposals are given to ANNOY library. Annoy assigns indices to each of the features. To retrieve nearest neighbor for any of the feature, it measures the Euclidean distance to all other features and indices of neighbors are assigned in the increasing order of their Euclidean distance. It has many advantages as compared to other nearest neighbor search algorithms. These include (i) small memory footprint (ii) Annoy creates data structures (read-only) which can be shared among multiple processes.
\subsection{Segmentation}
\label{sec:segment}
Segmentation is performed on the retrieved similar object proposals. We used Fully convolutional Networks for semantic segmentation proposed by Jonathan Long et al.\cite{long2015fully}. Convolutional networks are used as powerful tools for extracting the hierarchy of features. It was the first approach to generate pixel-wise predictions using supervised learning.  The contemporary classification networks are adapted to the segmentation tasks by transferring the learned representations. It utilizes a skip architecture which combines the semantic information from deep (coarse information) and shallow (fine appearance information) layers. %The fully connected (FC) layers are converted into fully convolutional layers to generate coarse output maps. The basic difference between FC layers and convolutional layers is that each neuron in convolutional layer is connected to a local region whereas neurons in FC layers are connected to all the activations in the previous layers. FC layers are converted into convolutional layers by using a filter of size equal to size of input volume to FC layer. 
The fully connected layers can also be considered as convolutions with kernels covering entire image. 
Transforming the FC layers into convolutional layers converts the classification network to generate a heat map. However, the generated output maps are of reduced size as compared to the input size. So, dense predictions are made from coarse maps by upsampling. Upsampling is performed by backward convolution (also called as deconvolution) with stride as $f$. Skip layers are added to fuse semantic and appearance information. 
\subsection{Visual Summary based on relative importance}
A visual summary is created from the segmented proposals. While retrieving the similar object proposals using ANNOY library, we preserved the Euclidean distances corresponding to each of the proposals. A basic collage is formed with 10 slots constituting the most similar proposal (least Euclidean distance) getting a larger block. The remaining segmented objects are placed in the other slots and a background is added to the image.
\section{Experimental Results}
\label{sec:results}
In this section, we discuss the empirical results on two publicly available benchmark co-segmentation datasets. We describe the datasets used followed by implementation details and baseline. Caffe \cite{jia2014caffe} is used for the constructing the Siamese network.
%\subsection{Experimental Setup}

\textbf{Datasets.} MSRC dataset \cite{shotton2006textonboost} consists of 14 categories. Each category consists of 30 images of dimension 213x320. iCoseg dataset \cite{batra2010icoseg} consists of 38 categories. Each category consists of about 20 to 30 images, which are of 300x500 size.

\textbf{Baselines and Parameter setting.} We report results with two baselines. The first baseline involves training the Siamese network with pretrained ILSVRC \cite{russakovsky2015imagenet} models. The weights are fine-tuned for target classes as in the datasets and then segmentation is performed on the clustered test set data. In the second baseline, we train the network on non-target classes and test the generalization ability on target classes. We evaluated on two objective measures: Precision ($\bar{\textit{P}}$) and Jaccard Similarity ($\bar{\textit{J}}$). $\bar{\textit{P}}$ indicates the fraction of the pixels in the segmented image common with the ground truth. $\bar{\textit{J}}$ is the intersection over union measure with the ground truth images.    
\begin{figure}[hbtp]
\centering
{
\includegraphics[scale=0.5]{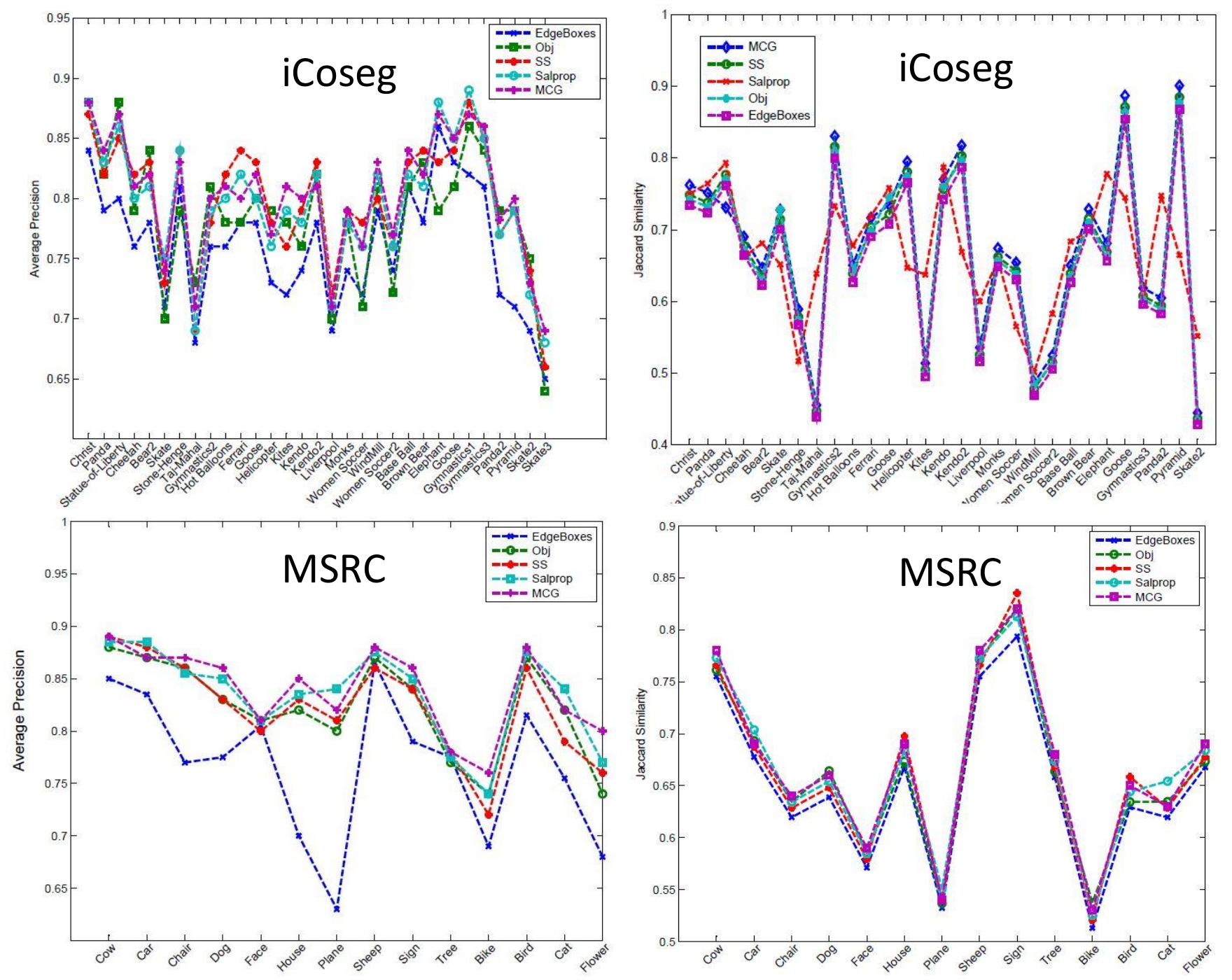}

\caption{Performance analysis of various object proposal generation methods with proposed architecture.}
\label{fig:objectprops}
}
\end{figure}

\begin{figure}[hbtp]
\centering
{
\includegraphics[scale=0.4]{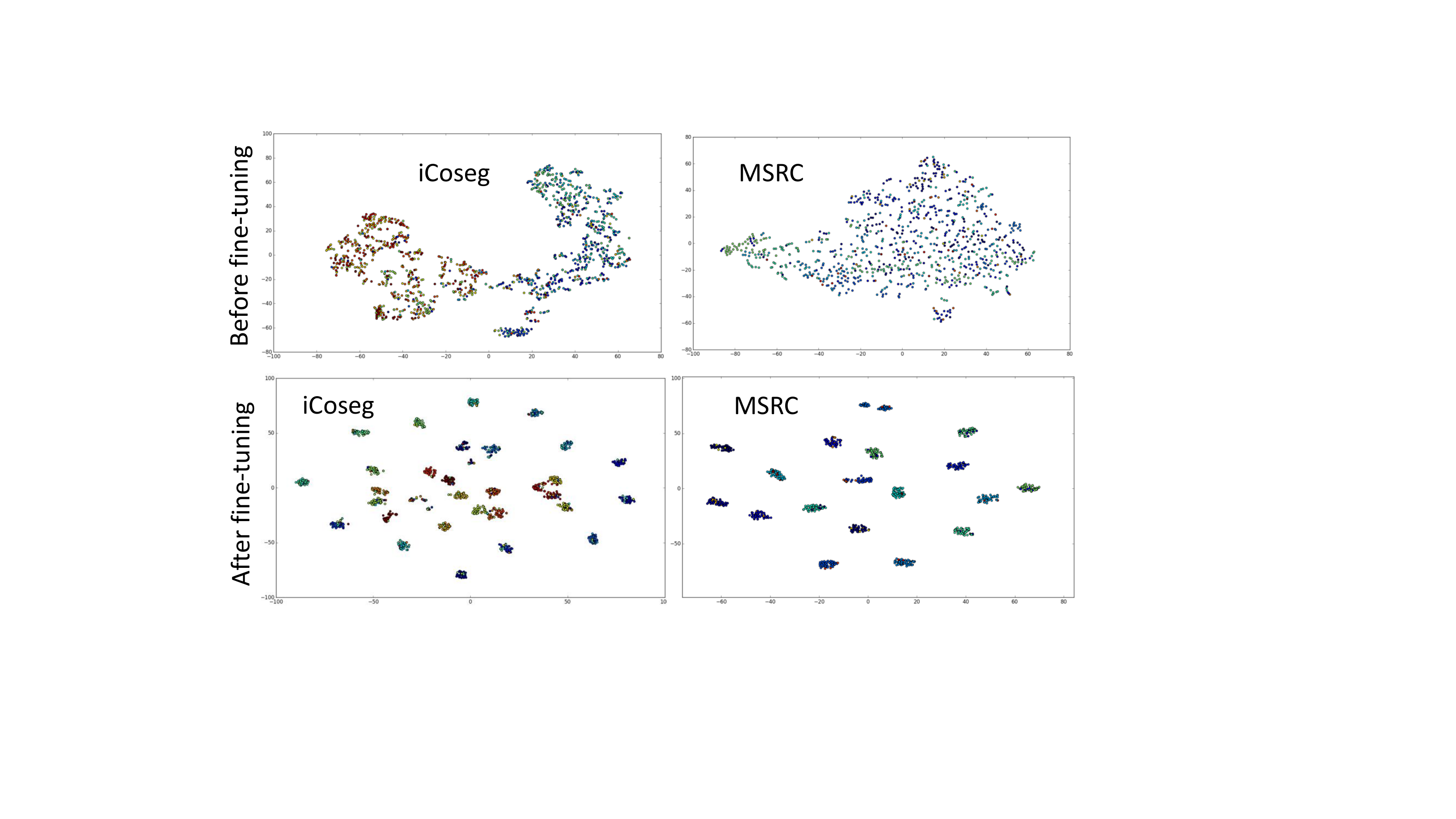}

\caption{Visualization of iCoseg and MSRC Training set using t-SNE}
\label{fig:tsne}
}
\end{figure}
We generated the object proposals using different methods and evaluated the performance on these metrics. The techniques used are Multiscale Combinatorial Grouping (MCG) \cite{pont2017multiscale}, Selective Search (SS) \cite{uijlings2013selective}, Objectness (Obj) \cite{alexe2012measuring}, SalProp \cite{mukherjee2017salprop} and Edgeboxes \cite{zitnick2014edge}. We further perform a non-maximal suppression and near duplicate rejection in the proposal set. We preserved the top-10 object proposals, so that all the object instances in the images are covered. We used GoogLeNet architectures \cite{szegedy2015going} for training the Siamese in our experiments. We used transfer learning, in which we initialized the weights with pre-trained model weights. We then fine-tuned the weights using back propagation technique. Siamese network is trained and the N-Dimensional (N=256) features are extracted for the test images. The N-Dimensional features are fed to ANNOY and similar object proposals are retrieved. The parameters used for Annoy library include number of trees, $ n_{trees}$=350 and number of nodes to inspect during searching $search_k$=50. Similar object proposals are segmented using FCN based semantic segmentation as discussed in Sec. \ref{sec:segment}. We trained the Siamese architecture by employing the standard backpropagation on feed-forward nets by stochastic gradient descent with momentum to adjust the weights. The mini-batch size was set to 128, with an equal learning rate for all layers set to 0.01. The number of iterations is set as 100,000, contrastive loss margin as 1. 

We also trained the Siamese network on datasets which contains similar (but not same) classes to iCoseg and MSRC datasets. We used Pascal \cite{everingham2010pascal}, Animals\cite{afkham2008joint} and Coseg-Rep \cite{dai2013cosegmentation} datasets to train the Siamese model and tested on iCoseg and MSRC datasets. Initially, we randomly selected positive and negative pairs for training the Siamese network. However, once most of the pairs are correctly learned, then using those pairs, Siamese cannot learn anymore. So, to address this issue, we used strategy of aggressive mining \cite{simo2015discriminative} for preparing hard negative and positive pairs. 

\textbf{Results.} We divided iCoseg dataset into 80\% training samples and 20\% as testing set for each class. For MSRC dataset the split was 70\%-30\% (training-test). The results of the $\bar{\textit{P}}$ and $\bar{\textit{J}}$ are shown in Fig.\ref{fig:objectprops}. It can be observed that  Siamese network fed with MCG proposals outperforms all other object proposal generation techniques with the closest being SalProp followed by SS, Obj and Edgeboxes. For both the datasets, the average precision and Jaccard index over all the classes with MCG proposals is higher than SalProp technique with a gap on an average being 2.48\% and 1.84\% in $\bar{\textit{P}}$ and $\bar{\textit{J}}$ respectively. The 256-D feature vector of the training set are visualized using t-SNE (t-Distributed Stochastic Neighbor Embedding) as shown in Fig. \ref{fig:tsne}. Firstly for high dimensional data, a probability distribution is built such that similar objects gets selected with high probability and dissimilar points have very low probability of being selected. In the second step, similar to a high-dimensional map, probability distribution over the points in the low-dimensional map is constructed. The color-coding in the t-SNE plots corresponds to the number of object classes in the respective datasets. Siamese net with post-processing helps in better separation of the classes compared to before fine-tuning. As can be seen, the results of clusters of classes are well separated with only few cluster of confusion. 

\begin{figure}[hbtp]
\centering
{
\includegraphics[scale=0.35]{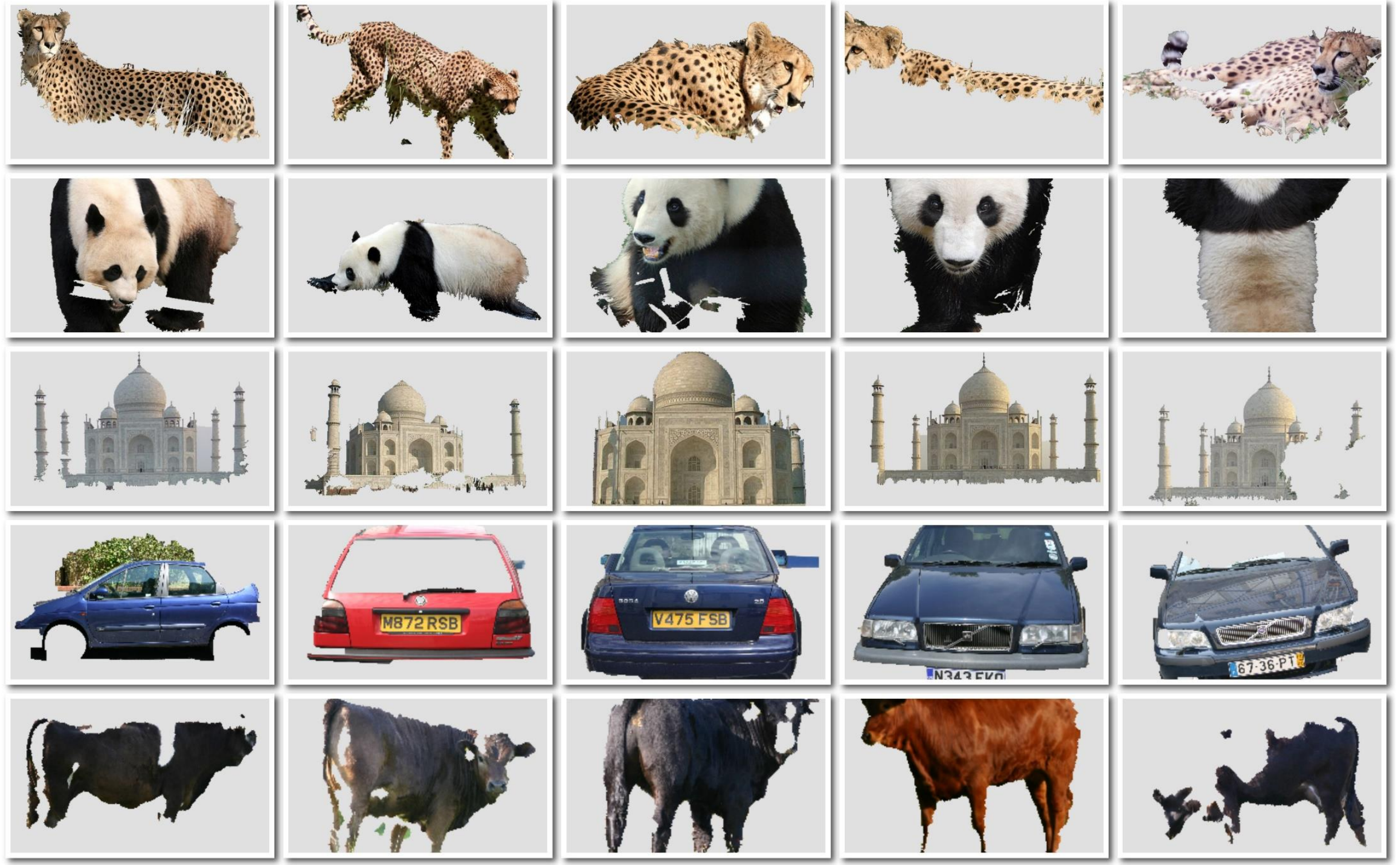}

\caption{Visual segmentation results on iCoseg and MSRC datasets. First three rows are classes of iCoseg (Cheetah, Panda, Taj-Mahal) and next two rows are MSRC (Car, Cow).}
\label{fig:segmented}
}
\end{figure}

We computed the average precision and jaccard similarity and compared with the other state-of-the-art methods in Tab. \ref{perficoseg}-\ref{perfmsrc}. On testing with complete iCoseg dataset, we achieve a gain in $\bar{\textit{P}}$ of 27.27\% (Joulin et al. \cite{joulin2010discriminative}), 23.52\% (Kim et al. \cite{kim2012multiple}). Quan et al. \cite{quan2016object} outperform with a margin of $\bar{\textit{P}}$ :9.67\% and $\bar{\textit{J}}$ : 13.15\% compared to the proposed technique (Siamese (MCG) + FCN segmentation). Similarly with MSRC dataset, we achieve a gain in $\bar{\textit{P}}$ of 20\% (Joulin et al. \cite{joulin2010discriminative}), 9.09\% (Jian et al. \cite{sun2016learning}), 44.82\% (Kim et al. \cite{kim2012multiple}) and in $\bar{\textit{J}}$ of 15.51\% (Yong Li\cite{li2016object}). Rubinstein et al.\cite{Rubinstein13Unsupervised} outperform with a margin of $\bar{\textit{P}}$ :8.6\% and $\bar{\textit{J}}$ : 1.47\% compared to the proposed technique (Siamese (MCG) + FCN segmentation). In FCN segmentation, we used VGGNet architecture, with FC layers replaced with convolutional layers. Deconvolutional layers are fixed using bilinear interpolation. We have abstained from using any auxiliary training and use the pretrained weights to avoid over-fitting in the FCN segmentation network. The object proposals that are similar and clustered together are fed as input to FCN segmentation to obtain the co-segmentation results. We consider only those object proposals for the cosegmentation task which have an intersection over union (IoU) score $IoU\geq0.5$. Since, the segmentation is performed on the tight object proposals it segments the regions specific to the object class only and thus refrains from performing semantic segmentation over entire image. Owing to the performance boost by aggressive mining we achieve an average gain of 4\% and 2.52\% on both the datasets in $\bar{\textit{P}}$ and $\bar{\textit{J}}$ respectively over training with MCG proposals (Siamese (MCG) + FCN segmentation). Fig. \ref{fig:segmented} shows the qualitative results on few example classes in the datasets. However, it is important to note here that since we perform cosegmentation on the subset of images owing to the retrieval results we observe that there is a drop in performance with respect to few reported techniques. The advantage of the proposed technique over other compared techniques(\cite{Rubinstein13Unsupervised,li2016object,joulin2010discriminative,kim2012multiple,quan2016object,meng2016cosegmentation,faktor2013co}) involve: (i) co-segmenting without explicit knowledge of localization prior in the form of saliency map (ii) co-segmentation pipeline being formulated as clustering followed by segmentation of the similar object classes thus eliminating the need for providing as input the relevant set of class-specific images as required in graph-based co-segmentation techniques. The proposed method takes less than 100ms for the generation of 256-dimensional features from trained Siamese network.

\begin{figure}[hbtp]
\centering
{
\includegraphics[scale=0.4]{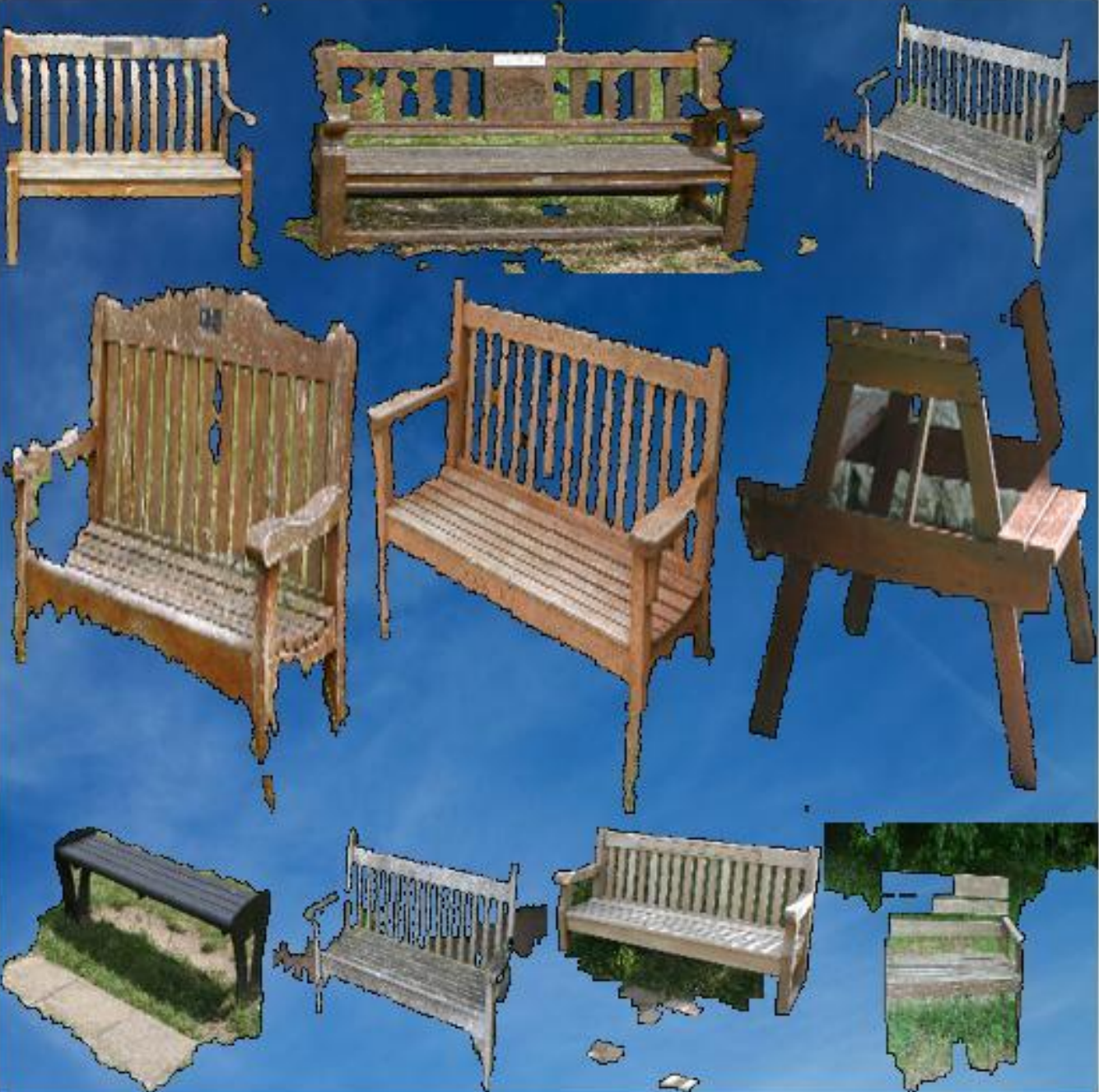}

\caption{Example of Collage results for Chair class (MSRC).}
\label{fig:collage}
}
\end{figure}

\begin{table}[]
\centering
\caption{Comparison of Average precision and Jaccard Similarity with state-of-the-art methods. ('-' indicates that the metric has not been provided in the respective paper) on iCoseg dataset}
\label{perficoseg}
\begin{tabular}{||c| c |c||}
\hline
\textbf{Method} & $\bar{\textit{P}}$ & $\bar{\textit{J}}$\\ \hline
\textbf{Rubinstein}\cite{Rubinstein13Unsupervised}                                       & 0.88                             & 0.674                            \\\hline
\textbf{Joulin}\cite{joulin2010discriminative}                                           & 0.66                             & -                                \\\hline
\textbf{Kim}\cite{kim2012multiple}                                              & 0.68                             & -                                \\\hline
\textbf{Keuttel}\cite{kuettel2012segmentation}                                          & 0.91                             & -                                \\\hline
\textbf{Quan}\cite{quan2016object}                                            & 0.93                             & 0.76                             \\\hline
\textbf{Fanman Meng}\cite{meng2016cosegmentation}                                      & -                                & 0.71                             \\\hline
\textbf{Faktor}\cite{faktor2013co}                                           & 0.92                             & 0.70                             \\\hline
%\textbf{Koteswar Rao}\cite{jerripothula2016image}                                     & -                                & 0.71                             \\\hline\hline
\multicolumn{3}{||c||}{\textbf{Trained on 80\% and tested with 20\%}}                                                                   \\\hline
\textbf{Method} & $\bar{\textit{P}}$ & $\bar{\textit{J}}$\\\hline
\textbf{Siamese (Edgeboxes) + FCN segmentation}                                   & 0.76                             & 0.61                             \\\hline
\textbf{Siamese (Obj) + FCN segmentation}                                         & 0.78                             & 0.62                             \\\hline
\textbf{Siamese (SS) + FCN segmentation}                                          & 0.79                             & 0.64                             \\\hline
\textbf{Siamese (SalProp) + FCN segmentation}                                     & 0.79                             & 0.64                             \\\hline
\textbf{Siamese (MCG) + FCN segmentation}                                         & 0.81                             & 0.654                            \\\hline\hline
\multicolumn{3}{||c||}{\textbf{Trained on 80\% and tested with 100\%}}   \\                                                  \hline             
\textbf{Method} & $\bar{\textit{P}}$ & $\bar{\textit{J}}$\\\hline
\textbf{Siamese (Edgeboxes) + FCN segmentation}                                   & 0.76                             & 0.62                             \\\hline
\textbf{Siamese (Obj) + FCN segmentation}                                         & 0.81                             & 0.64                             \\\hline
\textbf{Siamese (SS) + FCN segmentation}                                          & 0.83                             & 0.65                             \\\hline
\textbf{Siamese (SalProp) + FCN segmentation}                                     & 0.83                             & 0.65                             \\\hline
\textbf{Siamese (MCG) + FCN segmentation}                                         & 0.84                             & 0.66                             \\\hline\hline
\multicolumn{3}{||c||}{\textbf{\begin{tabular}[c]{@{}l@{}}Trained on Pascal+animals+coseg-rep and tested\\ on iCoseg\end{tabular}}} \\\hline
\textbf{Method} & $\bar{\textit{P}}$ & $\bar{\textit{J}}$\\\hline
\textbf{Siamese (MCG) + FCN segmentation}                                         & 0.73                             & 0.59                             \\\hline
\textbf{Siamese (MCG) + FCN segmentation-Aggressive mining}                       & 0.76                             & 0.61     \\              \hline         
\end{tabular}
\end{table}

\begin{table}[]
\centering
\caption{Comparision of Average precision and Jaccard Similarity with state-of-the-art methods. ('-' indicates that the metric has not been provided in the respective paper) on MSRC dataset}
\label{perfmsrc}
\begin{tabular}{||c| c |c||}
\hline
\textbf{Method} & $\bar{\textit{P}}$ & $\bar{\textit{J}}$\\ \hline
\textbf{Rubinstein}\cite{Rubinstein13Unsupervised}                                       & 0.92                            & 0.68                            \\\hline
\textbf{Joulin}\cite{joulin2010discriminative}                                           & 0.70                             & -                                \\\hline
\textbf{Jian Sun}\cite{sun2016learning}                                           & 0.77                             & 0.54                                \\\hline
\textbf{Faktor}\cite{faktor2013co}                                           & 0.89                             & 0.73                             \\\hline
\textbf{Kim}\cite{kim2012multiple}                                              & 0.58                             & -                                \\\hline
\textbf{Yong Li}\cite{li2016object}                                          & -                            & 0.58                                \\\hline\hline
 \multicolumn{3}{||c||}{\textbf{Trained on 70\% and tested with 30\%}}                                                                   \\\hline
\textbf{Method} & $\bar{\textit{P}}$ & $\bar{\textit{J}}$\\\hline
\textbf{Siamese (Edgeboxes) + FCN segmentation}                                   & 0.77                             & 0.62                             \\\hline
\textbf{Siamese (Obj) + FCN segmentation}                                         & 0.80                            & 0.63                             \\\hline
\textbf{Siamese (SS) + FCN segmentation}                                          & 0.81                             & 0.63                             \\\hline
\textbf{Siamese (SalProp) + FCN segmentation}                                     & 0.81                             & 0.64                             \\\hline
\textbf{Siamese (MCG) + FCN segmentation}                                         & 0.83                             & 0.65                            \\\hline\hline
\multicolumn{3}{||c||}{\textbf{Trained on 70\% and tested with 100\%}}   \\                                                  \hline             
\textbf{Method} & $\bar{\textit{P}}$ & $\bar{\textit{J}}$\\\hline
\textbf{Siamese (Edgeboxes) + FCN segmentation}                                   & 0.765                             & 0.64                             \\\hline
\textbf{Siamese (Obj) + FCN segmentation}                                         & 0.81                             & 0.65                             \\\hline
\textbf{Siamese (SS) + FCN segmentation}                                          & 0.82                             & 0.65                             \\\hline
\textbf{Siamese (SalProp) + FCN segmentation}                                     & 0.82                             & 0.66                             \\\hline
\textbf{Siamese (MCG) + FCN segmentation}                                         & 0.84                             & 0.67                            \\\hline\hline
\multicolumn{3}{||c||}{\textbf{\begin{tabular}[c]{@{}l@{}}Trained on Pascal+animals+coseg-rep and tested\\ on iCoseg\end{tabular}}} \\\hline
\textbf{Method} & $\bar{\textit{P}}$ & $\bar{\textit{J}}$\\\hline
\textbf{Siamese (MCG) + FCN segmentation}                                         & 0.76                             & 0.60                             \\\hline
\textbf{Siamese (MCG) + FCN segmentation-Aggressive mining}                       & 0.79                             & 0.61     \\              \hline         
\end{tabular}
\end{table}
We create a visual summary of the co-segmented similar objects. We preserved the Euclidean distances while retrieving the similar objects. Image is divided into different blocks and objects are placed such that the object with least Euclidean distance is at the center. A proper background is added to improve the visual appearance. Fig. \ref{fig:collage} shows the sample collage results formed the Chair class in MSRC. A 512x512 image is divided into 10 blocks consisting a blue sky back-ground to form collage. Future work would be aimed to further improve the segmentation results and utilization of more cues for the relevance ranking in the collage-generation.

\section{Conclusion} \label{sec:conclusion}
We addressed object cosegmentation and posed it as a clustering problem using deep Siamese network to align the similar images which are segmented using semantic segmentation. Additionally, we compared the performance of various object proposal generation schemes on Siamese architecture. We performed extensive evaluation on iCoseg and MSRC dataset and demonstrated that the deep features can encode the commonness prior and thus provide a more discriminative representation for the features.  

\bibliographystyle{IEEEtran}
\bibliography{sigproc}

\begin{thebibliography}{10}
\providecommand{\url}[1]{#1}
\csname url@samestyle\endcsname
\providecommand{\newblock}{\relax}
\providecommand{\bibinfo}[2]{#2}
\providecommand{\BIBentrySTDinterwordspacing}{\spaceskip=0pt\relax}
\providecommand{\BIBentryALTinterwordstretchfactor}{4}
\providecommand{\BIBentryALTinterwordspacing}{\spaceskip=\fontdimen2\font plus
\BIBentryALTinterwordstretchfactor\fontdimen3\font minus
  \fontdimen4\font\relax}
\providecommand{\BIBforeignlanguage}[2]{{%
\expandafter\ifx\csname l@#1\endcsname\relax
\typeout{** WARNING: IEEEtran.bst: No hyphenation pattern has been}%
\typeout{** loaded for the language `#1'. Using the pattern for}%
\typeout{** the default language instead.}%
\else
\language=\csname l@#1\endcsname
\fi
#2}}
\providecommand{\BIBdecl}{\relax}
\BIBdecl

\bibitem{vicente2011object}
S.~Vicente, C.~Rother, and V.~Kolmogorov, ``Object cosegmentation,'' in
  \emph{Computer Vision and Pattern Recognition (CVPR), 2011 IEEE Conference
  on}.\hskip 1em plus 0.5em minus 0.4em\relax IEEE, 2011, pp. 2217--2224.

\bibitem{wang2015robust}
W.~Wang, J.~Shen, X.~Li, and F.~Porikli, ``Robust video object
  cosegmentation,'' \emph{Image Processing, IEEE Transactions on}, vol.~24,
  no.~10, pp. 3137--3148, 2015.

\bibitem{li2016object}
Y.~Li, J.~Liu, Z.~Li, H.~Lu, and S.~Ma, ``Object co-segmentation via salient
  and common regions discovery,'' \emph{Neurocomputing}, vol. 172, pp.
  225--234, 2016.

\bibitem{Huang2016ObjectCB}
L.~Huang, R.~Gan, and G.~Zeng, ``Object cosegmentation by similarity
  propagation with saliency information and objectness frequency map,''
  \emph{2016 3rd International Conference on Systems and Informatics (ICSAI)},
  pp. 906--911, 2016.

\bibitem{rother2006cosegmentation}
C.~Rother, T.~Minka, A.~Blake, and V.~Kolmogorov, ``Cosegmentation of image
  pairs by histogram matching-incorporating a global constraint into mrfs,'' in
  \emph{Computer Vision and Pattern Recognition, 2006 IEEE Computer Society
  Conference on}, vol.~1.\hskip 1em plus 0.5em minus 0.4em\relax IEEE, 2006,
  pp. 993--1000.

\bibitem{wang2013image}
F.~Wang, Q.~Huang, and L.~Guibas, ``Image co-segmentation via consistent
  functional maps,'' in \emph{Proceedings of the IEEE International Conference
  on Computer Vision}, 2013, pp. 849--856.

\bibitem{joulin2012multi}
A.~Joulin, F.~Bach, and J.~Ponce, ``Multi-class cosegmentation,'' in
  \emph{Computer Vision and Pattern Recognition (CVPR), 2012 IEEE Conference
  on}.\hskip 1em plus 0.5em minus 0.4em\relax IEEE, 2012, pp. 542--549.

\bibitem{batra2010icoseg}
D.~Batra, A.~Kowdle, D.~Parikh, J.~Luo, and T.~Chen, ``icoseg: Interactive
  co-segmentation with intelligent scribble guidance,'' in \emph{Computer
  Vision and Pattern Recognition (CVPR), 2010 IEEE Conference on}.\hskip 1em
  plus 0.5em minus 0.4em\relax IEEE, 2010, pp. 3169--3176.

\bibitem{kowdle2010imodel}
A.~Kowdle, D.~Batra, W.-C. Chen, and T.~Chen, ``imodel: Interactive
  co-segmentation for object of interest 3d modeling,'' in \emph{Trends and
  Topics in Computer Vision}.\hskip 1em plus 0.5em minus 0.4em\relax Springer,
  2010, pp. 211--224.

\bibitem{mukherjee2009half}
L.~Mukherjee, V.~Singh, and C.~R. Dyer, ``Half-integrality based algorithms for
  cosegmentation of images,'' in \emph{Computer Vision and Pattern Recognition,
  2009. CVPR 2009. IEEE Conference on}.\hskip 1em plus 0.5em minus 0.4em\relax
  IEEE, 2009, pp. 2028--2035.

\bibitem{long2015fully}
J.~Long, E.~Shelhamer, and T.~Darrell, ``Fully convolutional networks for
  semantic segmentation,'' in \emph{Proceedings of the IEEE Conference on
  Computer Vision and Pattern Recognition}, 2015, pp. 3431--3440.

\bibitem{jia2014caffe}
Y.~Jia, E.~Shelhamer, J.~Donahue, S.~Karayev, J.~Long, R.~Girshick,
  S.~Guadarrama, and T.~Darrell, ``Caffe: Convolutional architecture for fast
  feature embedding,'' \emph{arXiv preprint arXiv:1408.5093}, 2014.

\bibitem{shotton2006textonboost}
J.~Shotton, J.~Winn, C.~Rother, and A.~Criminisi, ``Textonboost: Joint
  appearance, shape and context modeling for multi-class object recognition and
  segmentation,'' in \emph{European conference on computer vision}.\hskip 1em
  plus 0.5em minus 0.4em\relax Springer, 2006, pp. 1--15.

\bibitem{russakovsky2015imagenet}
O.~Russakovsky, J.~Deng, H.~Su, J.~Krause, S.~Satheesh, S.~Ma, Z.~Huang,
  A.~Karpathy, A.~Khosla, M.~Bernstein \emph{et~al.}, ``Imagenet large scale
  visual recognition challenge,'' \emph{International Journal of Computer
  Vision}, vol. 115, no.~3, pp. 211--252, 2015.

\bibitem{pont2017multiscale}
J.~Pont-Tuset, P.~Arbelaez, J.~T. Barron, F.~Marques, and J.~Malik,
  ``Multiscale combinatorial grouping for image segmentation and object
  proposal generation,'' \emph{IEEE transactions on pattern analysis and
  machine intelligence}, vol.~39, no.~1, pp. 128--140, 2017.

\bibitem{uijlings2013selective}
J.~R. Uijlings, K.~E. Van De~Sande, T.~Gevers, and A.~W. Smeulders, ``Selective
  search for object recognition,'' \emph{International journal of computer
  vision}, vol. 104, no.~2, pp. 154--171, 2013.

\bibitem{alexe2012measuring}
B.~Alexe, T.~Deselaers, and V.~Ferrari, ``Measuring the objectness of image
  windows,'' \emph{IEEE transactions on pattern analysis and machine
  intelligence}, vol.~34, no.~11, pp. 2189--2202, 2012.

\bibitem{mukherjee2017salprop}
P.~Mukherjee, B.~Lall, and S.~Tandon, ``Salprop: Salient object proposals via
  aggregated edge cues,'' \emph{arXiv preprint arXiv:1706.04472}, 2017.

\bibitem{zitnick2014edge}
C.~L. Zitnick and P.~Doll{\'a}r, ``Edge boxes: Locating object proposals from
  edges.'' in \emph{ECCV (5)}, 2014, pp. 391--405.

\bibitem{szegedy2015going}
C.~Szegedy, W.~Liu, Y.~Jia, P.~Sermanet, S.~Reed, D.~Anguelov, D.~Erhan,
  V.~Vanhoucke, and A.~Rabinovich, ``Going deeper with convolutions,'' in
  \emph{Proceedings of the IEEE Conference on Computer Vision and Pattern
  Recognition}, 2015, pp. 1--9.

\bibitem{everingham2010pascal}
M.~Everingham, L.~Van~Gool, C.~K. Williams, J.~Winn, and A.~Zisserman, ``The
  pascal visual object classes (voc) challenge,'' \emph{International journal
  of computer vision}, vol.~88, no.~2, pp. 303--338, 2010.

\bibitem{afkham2008joint}
H.~M. Afkham, A.~T. Targhi, J.-O. Eklundh, and A.~Pronobis, ``Joint visual
  vocabulary for animal classification,'' in \emph{Pattern Recognition, 2008.
  ICPR 2008. 19th International Conference on}.\hskip 1em plus 0.5em minus
  0.4em\relax IEEE, 2008, pp. 1--4.

\bibitem{dai2013cosegmentation}
J.~Dai, Y.~Nian~Wu, J.~Zhou, and S.-C. Zhu, ``Cosegmentation and cosketch by
  unsupervised learning,'' in \emph{Proceedings of the IEEE international
  conference on computer vision}, 2013, pp. 1305--1312.

\bibitem{simo2015discriminative}
E.~Simo-Serra, E.~Trulls, L.~Ferraz, I.~Kokkinos, P.~Fua, and F.~Moreno-Noguer,
  ``Discriminative learning of deep convolutional feature point descriptors,''
  in \emph{Proceedings of the IEEE International Conference on Computer
  Vision}, 2015, pp. 118--126.

\bibitem{joulin2010discriminative}
A.~Joulin, F.~Bach, and J.~Ponce, ``Discriminative clustering for image
  co-segmentation,'' in \emph{Computer Vision and Pattern Recognition (CVPR),
  2010 IEEE Conference on}.\hskip 1em plus 0.5em minus 0.4em\relax IEEE, 2010,
  pp. 1943--1950.

\bibitem{kim2012multiple}
G.~Kim and E.~P. Xing, ``On multiple foreground cosegmentation,'' in
  \emph{Computer Vision and Pattern Recognition (CVPR), 2012 IEEE Conference
  on}.\hskip 1em plus 0.5em minus 0.4em\relax IEEE, 2012, pp. 837--844.

\bibitem{quan2016object}
R.~Quan, J.~Han, D.~Zhang, and F.~Nie, ``Object co-segmentation via graph
  optimized-flexible manifold ranking,'' in \emph{Proceedings of the IEEE
  Conference on Computer Vision and Pattern Recognition}, 2016, pp. 687--695.

\bibitem{sun2016learning}
J.~Sun and J.~Ponce, ``Learning dictionary of discriminative part detectors for
  image categorization and cosegmentation,'' \emph{International Journal of
  Computer Vision}, vol. 120, no.~2, pp. 111--133, 2016.

\bibitem{Rubinstein13Unsupervised}
M.~Rubinstein, A.~Joulin, J.~Kopf, and C.~Liu, ``Unsupervised joint object
  discovery and segmentation in internet images,'' \emph{IEEE Conf. on Computer
  Vision and Pattern Recognition (CVPR)}, June 2013.

\bibitem{meng2016cosegmentation}
F.~Meng, J.~Cai, and H.~Li, ``Cosegmentation of multiple image groups,''
  \emph{Computer Vision and Image Understanding}, vol. 146, pp. 67--76, 2016.

\bibitem{faktor2013co}
A.~Faktor and M.~Irani, ``Co-segmentation by composition,'' in
  \emph{Proceedings of the IEEE International Conference on Computer Vision},
  2013, pp. 1297--1304.

\bibitem{kuettel2012segmentation}
D.~Kuettel, M.~Guillaumin, and V.~Ferrari, ``Segmentation propagation in
  imagenet,'' \emph{Computer Vision--ECCV 2012}, pp. 459--473, 2012.

\end{thebibliography}

\end{document}